\documentclass[runningheads]{llncs}

\usepackage{eccv}

\usepackage{eccvabbrv}

\usepackage{graphicx}
\usepackage{booktabs}

\usepackage[accsupp]{axessibility}  

\usepackage{hyperref}

\usepackage{orcidlink}

\begin{document}

\title{VividDreamer: Invariant Score Distillation For Hyper-Realistic Text-to-3D Generation}

\titlerunning{VividDreamer}

\author{Wenjie Zhuo\inst{1,2}\orcidlink{0009-0002-0851-5546} \and
Fan Ma\inst{2}\orcidlink{0000-0002-4131-1222} \and
Hehe Fan\inst{2}\orcidlink{0000-0001-9572-2345} \and 
Yi Yang\inst{1,2\dagger}\orcidlink{0000-0002-0512-880X}}

\authorrunning{W. Zhuo, F. Ma et al.}
\institute{State Key Laboratory of Brain-machine Intelligence, Zhejiang University \and ReLER Lab, CCAI, Zhejiang University \\
\email{12021057@zju.edu.cn} \\
}
\renewcommand{\thefootnote}{\fnsymbol{footnote}}
\footnotetext{†Corresponding author.}
\footnotetext{Our code is available at \url{https://github.com/SupstarZh/VividDreamer}.}
\renewcommand{\thefootnote}{\arabic{footnote}}

\maketitle
\begin{abstract}
This paper presents Invariant Score Distillation (ISD), a novel method for high-fidelity text-to-3D generation. ISD aims to tackle the over-saturation and over-smoothing problems in Score Distillation Sampling (SDS). In this paper, SDS is decoupled into a weighted sum of two components: the reconstruction term and the classifier-free guidance term.  We experimentally found that over-saturation stems from the large classifier-free guidance scale and over-smoothing comes from the reconstruction term. To overcome these problems, ISD utilizes an invariant score term derived from DDIM sampling to replace the reconstruction term in SDS. This operation allows the utilization of a medium classifier-free guidance scale and mitigates the reconstruction-related errors, thus preventing the over-smoothing and over-saturation of results. Extensive experiments demonstrate that our method greatly enhances SDS and produces realistic 3D objects through single-stage optimization. 
\end{abstract}

\begin{figure}[!ht]
    \centering
    \includegraphics[width=0.98\textwidth]{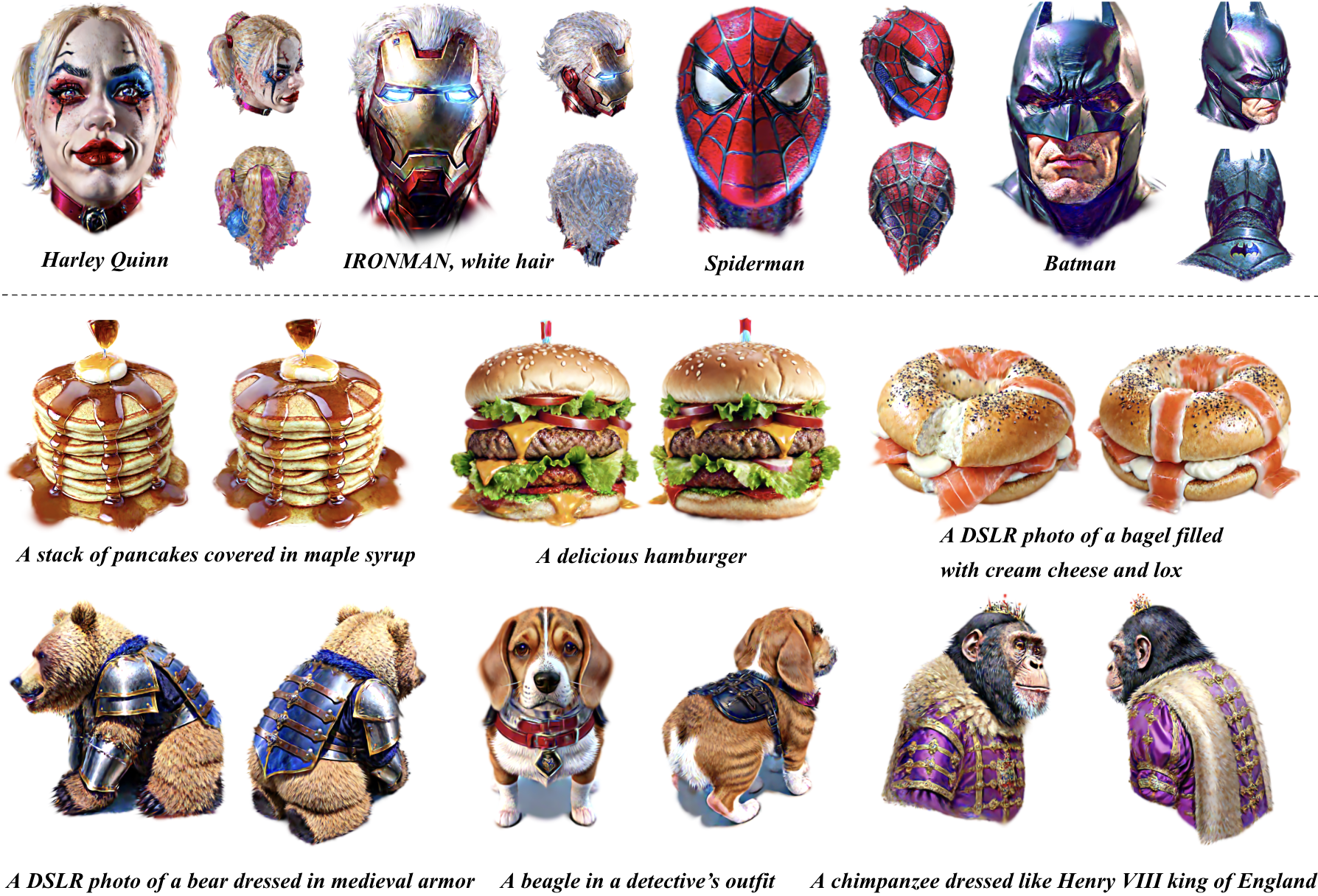}
    \caption{
        Examples generated by our framework. Our methods can generate detailed and high-fidelity 3D objects from a wide range of textual prompts.
    }
    \label{fig:main_fig}
\end{figure}

\section{Introduction}
This paper considers text-to-3D content generation \cite{jun2023shap,wang2024prolificdreamer,poole2022dreamfusion,nichol2022point,xu2023seeavatar,chen2024survey}. It is a fundamental task in computer vision with the potential to benefit various downstream applications. Benefiting from large-scale text-image pairs, there recently emerges a new 3D generation method, called 2D lifting methods. The main idea of these methods is to utilize text-to-image generative models \cite{song2020score,song2020denoising,ho2020denoising,radford2021learning,li2022blip} to guide the 3D creation. Early methods like DreamFields \cite{jain2021dreamfields} and CLIP-forge \cite{sanghi2022clipforge} use the CLIP \cite{radford2021learning,li2022blip} as guidance. Recently, inspired by the tremendous success of diffusion models \cite{song2020score,song2020denoising,ho2020denoising} in the realm of image generation and conditional image generation \cite{nichol2021glide,saharia2022photorealistic,rombach2022high,yang2024doraemongpt}, DreamFusion \cite{poole2022dreamfusion} introduces a loss based on probabilistic density distillation \cite{oord2017parallel}, called Score Distillation Sampling (SDS). They utilize pre-trained diffusion priors to optimize 3D representations \cite{mildenhall2020nerf,mueller2022instant,kerbl3Dgaussians} by steering their rendered images toward regions with higher probability density, conditioned on the text. This technology has sparked an abundance of subsequent research efforts. For example, Liu \etal \cite{liu2023zero1to3} propose Zero123, which introduces SDS into the single-image reconstruction task \cite{qian2023magic123,liu2024one,liu2023syncdreamer}. However, despite its widespread impact, the 3D objects generated by SDS frequently suffer from over-saturation and over-smoothing problems\cite{wang2024prolificdreamer,armandpour2023re,zhu2023hifa}.

In this paper, we take an in-depth analysis of these issues. We decompose the SDS into two components, one is the reconstruction term: $\epsilon_\phi(z_t;y,t)-\epsilon$, and the other is the classifier-free guidance term: $\epsilon_\phi(z_t;y,t)-\epsilon_\phi(z_t;\varnothing, t)$. We use each term individually as loss for text-to-image generation, and find two problems: 1) using the reconstruction term alone not only lacks semantic alignment functionality but also deteriorates the originally high-quality images. 2) using the classifier-free guidance term alone suffers from serious over-saturation (as shown in Fig. \ref{fig:items}). SDS avoids problem 1 by assigning a large guidance scale (100.0) to the classifier-free guidance term, but this exacerbates problem 2.

To overcome these problems, we further analyze the causes of problem 1 and focus on optimizing the reconstruction term. We find the function of this item is actually to reconstruct the noise image in single-step, and this leads to problem 1 from two aspects: 1) At the beginning of the generation, reconstructing semantic-free images hinders semantic alignment. 2) When the image is aligned with the condition, the inconsistency of the single-step reconstruction destroys the image and causes over-smoothing. To address these, we propose the Invariant Score Distillation (ISD) method, which replaces the reconstruction item with the invariant score term. This term no longer reconstructs the noisy image, but aligns the noisy image to the previous time step state. This brings two benefits: Firstly, the invariant score prior can be continuously refined during the generation process to provide more details for the results while avoiding errors caused by single-step reconstruction, avoiding over-smoothing and details lost problems. Secondly, the introduction of invariant score prior allows us to use a conventional guidance scale (7.5) for text-to-3D generation, avoiding the over-saturation problem. We also conduct extensive experiments to verify the effectiveness of our method. The results show that our method can generate realistic and detailed 3D objects. Compared with existing methods, our method shows strong competitiveness both qualitatively and quantitatively. 

Overall, our contributions can be summarized as follows:

(1) We decompose the SDS into a classifier-free guidance term and a reconstruction term, and identify the over-saturation comes from the large classifier-free guidance scale, and the over-smoothing stems from the reconstruction term.

(2) We propose a novel Invariant Score Distillation (ISD) to overcome the aforementioned issues. ISD replaces the reconstruction term with an invariant score prior derived from DDIM, avoiding the inherent problems of SDS.

(3) We verify the effectiveness of our method through quantitative and qualitative experiments. The results show that our method can generate high-fidelity 3D objects in single-stage optimization and is highly competitive compared with existing methods.
\section{Related Work}
\subsection{Text to image Generation}
Recently, significant progress has been made in text-to-image generation \cite{rombach2022high, radford2021learning, zhuo2023whitenedcse, li2022blip,zhou2024migc}. Among these, diffusion models \cite{song2020denoising, song2020score, ho2020denoising, karras2022elucidating, lu2022dpm} are particularly notable for their outstanding performance. These models, once trained, are capable of transforming Gaussian noise into high-definition, realistic samples through multiple iterations. To enable conditional image generation, Dhariwal \etal \cite{dhariwal2021diffusion} introduces classifier guidance. This approach requires training an additional noisy image classifier, thereby needing extra gradient guidance during sampling, and limiting generated image variety. To address these issues, Ho \etal proposes classifier-free guidance \cite{ho2022classifier}, eliminating the need for an explicit classifier. This method trains the diffusion model in both conditioned and unconditioned modes, it modifies the model by Eq. \ref{eq:cfg}:
\begin{equation}
    \label{eq:cfg}
    \begin{aligned}
        \tilde{\epsilon}_\phi(z_t; y, t) = (1 + w)\epsilon_\phi(z_t; y, t) - w\epsilon_\phi(z_t; y=\varnothing, t)
    \end{aligned}
\end{equation}
where $\varnothing$ is an empty text prompt representing the unconditional case, and $w$ is a guidance scale, used to control the influence of conditional information. Building upon this methodology, the introduction of models such as GLIDE \cite{nichol2021glide}, Imagen \cite{saharia2022photorealistic}, and Stable Diffusion \cite{rombach2022high} have propelled diffusion models to excel in the text-to-image generation domain. Recently, the influence of these models has also extended to text-to-3D generation.

\subsection{Text to 3D Generation}
Text-to-3D generation refers to creating 3D assets based on text descriptions. Current works can be categorized into two main types: (i) Native 3D methods that directly train a new diffusion model with 3D data \cite{nichol2022point,hong2023lrm,jun2023shap,li2023instant3d,zeng2022lion} and (ii) 2D lifting methods that use the pre-trained 2D text-to-image diffusion models to guide the 3D assets generation \cite{poole2022dreamfusion,wang2023score,wang2024prolificdreamer,zhu2023hifa,huang2023dreamtime,yu2023text,katzir2023noise,liang2023luciddreamer,wu2024consistent3d,lin2023magic3d,chen2023fantasia3d,zhou2024headstudio,ye2024dreamreward,wang2023animatabledreamer}. Upon completion of pre-training, the former can rapidly generate 3D-consistent assets. However, due to the scarcity and expense of 3D datasets \cite{shapenet2015,deitke2023objaverse,deitke2024objaverse}, the performance and diversity of current 3D diffusion models are significantly constrained.

2D lifting methods leverage models pre-trained on rich 2D text-image datasets to guide the generation of 3D assets \cite{mueller2022instant,mildenhall2020nerf,kerbl3Dgaussians}. Dreamfusion \cite{poole2022dreamfusion} proposes Score Distillation Sampling (SDS) to utilize diffusion models for guiding the generation of 3D assets. Despite the popularity of this approach, empirical observations reveal that SDS often suffers from over-saturation and over-smoothing issues \cite{wang2024prolificdreamer,zhu2023hifa,huang2023dreamtime,yu2023text,lin2023magic3d,chen2023fantasia3d}. To address these challenges, Magic3D \cite{lin2023magic3d} adopts a two-stage method by combining NeRF \cite{mildenhall2020nerf} with mesh fine-tuning, Fantasia3D \cite{chen2023fantasia3d} takes a unique approach by decoupling geometry and appearance, facilitating the generation of explicit meshes and textures. Additionally, some methods refine the loss function to promote the quality of created 3D models. HiFA \cite{zhu2023hifa} introduces an image-level loss term based on the original SDS. ProlificDreamer \cite{wang2024prolificdreamer} proposes VSD, it replaces the random Gaussian noise with a learned score function of noisy rendered images conditioned on the camera pose $c$.  NFSD \cite{katzir2023noise} decomposes components of SDS and provides experimental improvements to the loss function, while CSD \cite{yu2023text} analyzes the components of the SDS and directly uses the classifier score term as guidance. Recently, with the rise of the new rendering technology 3DGS \cite{kerbl3Dgaussians}, a new batch of excellent works has emerged \cite{liang2023luciddreamer,tang2023dreamgaussian,yi2023gaussiandreamer,chen2023textto3d}. 
\section{Preliminaries}
\subsection{Score Distillation Sampling}
\label{sec:sds}
Score Distillation Sampling (SDS) \cite{poole2022dreamfusion} utilizes a pre-trained text-to-image diffusion model $\phi$ to guide the 3D representation parameterized by $\theta$. Specifically, for a given camera pose $\pi$, we denote $x=g(\theta; \pi)$ as the image rendered from a differentiable rendering function $g$. SDS ensures that the images obtained from any viewpoint via the differentiable rendering function $g$ are aligned with the textual prompt $y$. The form of SDS loss is as follows:
\begin{equation}
    \label{eq:sds_1}
    \begin{aligned}
       \nabla_\theta \mathcal{L}_{SDS} = \mathbf{E}_{t, \epsilon, \pi}[w(t)(\epsilon_\phi(z_t; y, t)-\epsilon)\frac{\partial x}{\partial \theta}]
    \end{aligned}
\end{equation}
where $w(t)$ is a weighting function, and $z_t$ represents a noisy version of $x$. In practice, SDS employs Classifier-free guidance \cite{ho2022classifier}, therefore, the $\epsilon_\phi(z_t; y, t)$ in Eq. \ref{eq:sds_1} is $\epsilon_\phi(z_t; y, t) + w(\epsilon_\phi(z_t; y, t)-\epsilon_\phi(z_t; \varnothing, t))$, where $w$ represents the guidance scale. Combine them, the final form of SDS is:

\begin{figure}[t]
    \centering
    \includegraphics[width=0.85\textwidth]{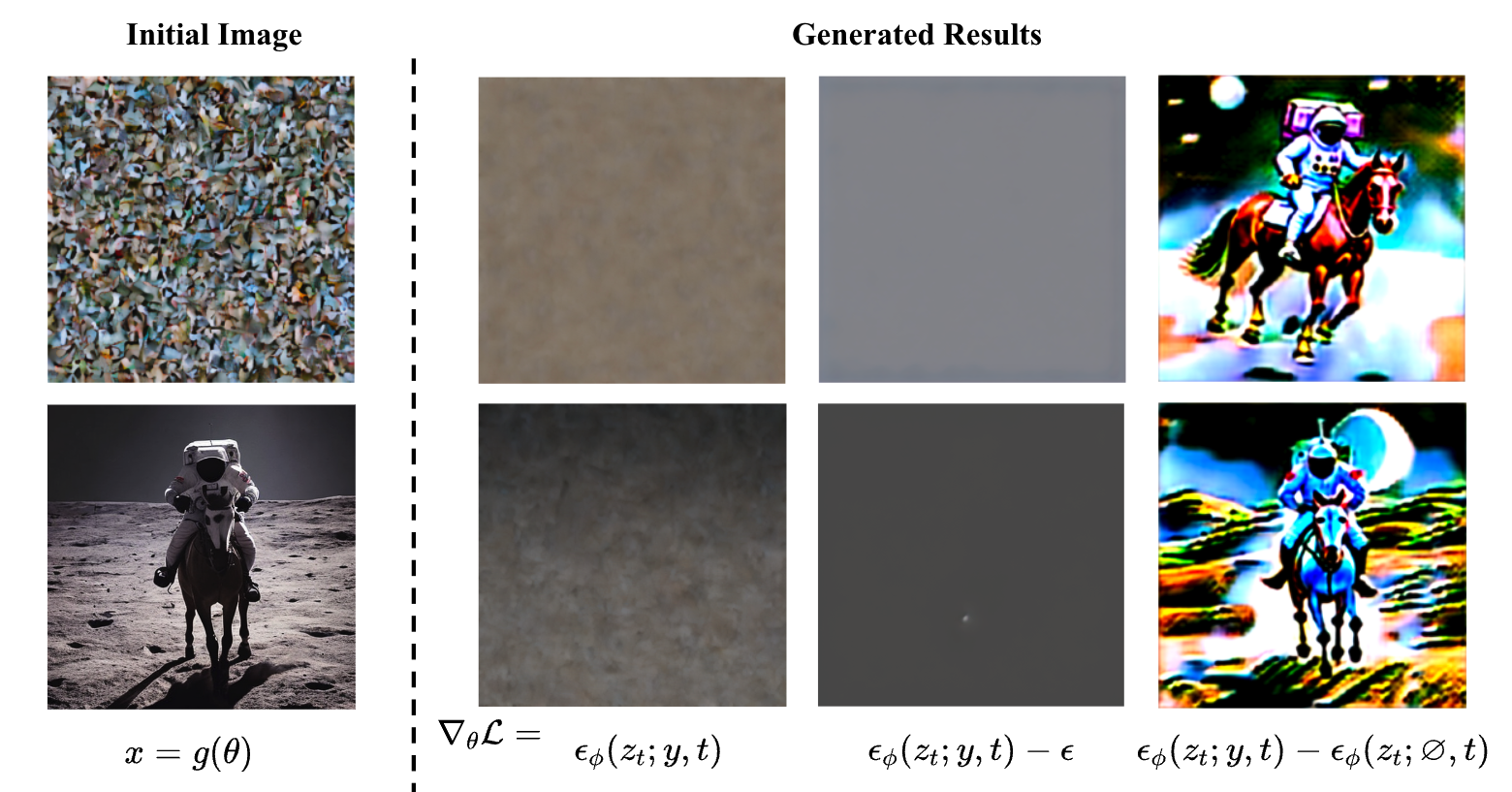}
    \caption{
        We separately utilize each term as the loss function to generate images in 2D experiments using SDS, with the prompt "An astronaut riding a horse". In the first line, we use a randomly noise as initialization, in the second line, we use a high-quality image sampled from Stable Diffusion 2-1 base \cite{rombach2022high} as initialization. The results show that regardless of the initialization method, using each term alone cannot generate realistic and detailed samples.
    }
    \label{fig:items}
\end{figure}

\begin{equation}
    \label{eq:sds}
    \begin{aligned}
       \nabla_\theta \mathcal{L}_{SDS} = \mathbf{E}_{t, \epsilon, \pi}[w(t)(\epsilon_\phi(z_t; y, t) + w(\epsilon_\phi(z_t; y, t)-\epsilon_\phi(z_t; \varnothing, t)) - \epsilon)\frac{\partial x}{\partial \theta}]
    \end{aligned}
\end{equation}
In this form, SDS consists of two parts. The first part is the original $\epsilon_\phi(z_t; y, t) - \epsilon$, which is proportional to $\nabla_{z_t}\log p(z_t|y)$. Its role is to restore $z_t$ back to its pre-noised state $x$ according to the condition $y$. The second part is introduced by Classifier-free guidance \cite{ho2022classifier}, $\epsilon_\phi(z_t; y, t)-\epsilon_\phi(z_t; \varnothing, t)$, which is proportional to $\nabla_{z_t}\log p(z_t|y)-\nabla_{z_t}\log p(z_t)$. According to Bayes' rule, this term can further be proportional to $\nabla_{z_t}\log p(y|z_t)$. It acts as an implicit noisy image classifier.



\section{Methodology}

\subsection{Anatomy of SDS}
\label{sec:deep_in_sds}
Based on the analysis in Sec. \ref{sec:sds}, we seek to explore why SDS suffers from over-smoothing and over-saturation. To this end, we conduct an experiment to investigate the role of each term in the generation process of SDS. Specifically, following our previous analysis, we divide the SDS loss into two components: one is the classifier-free guidance term $\epsilon_\phi(z_t; y, t)-\epsilon_\phi(z_t; \varnothing, t)$, and the other is the reconstruction term $\epsilon_\phi(z_t; y, t) - \epsilon$. We use a randomly generated noise and a high-quality image sampled from Stable Diffusion \cite{rombach2022high} as our initialization images, respectively. Then we use each term as the loss function for image generation. 

\begin{figure}[t]
    \centering
    \includegraphics[width=0.98\textwidth]{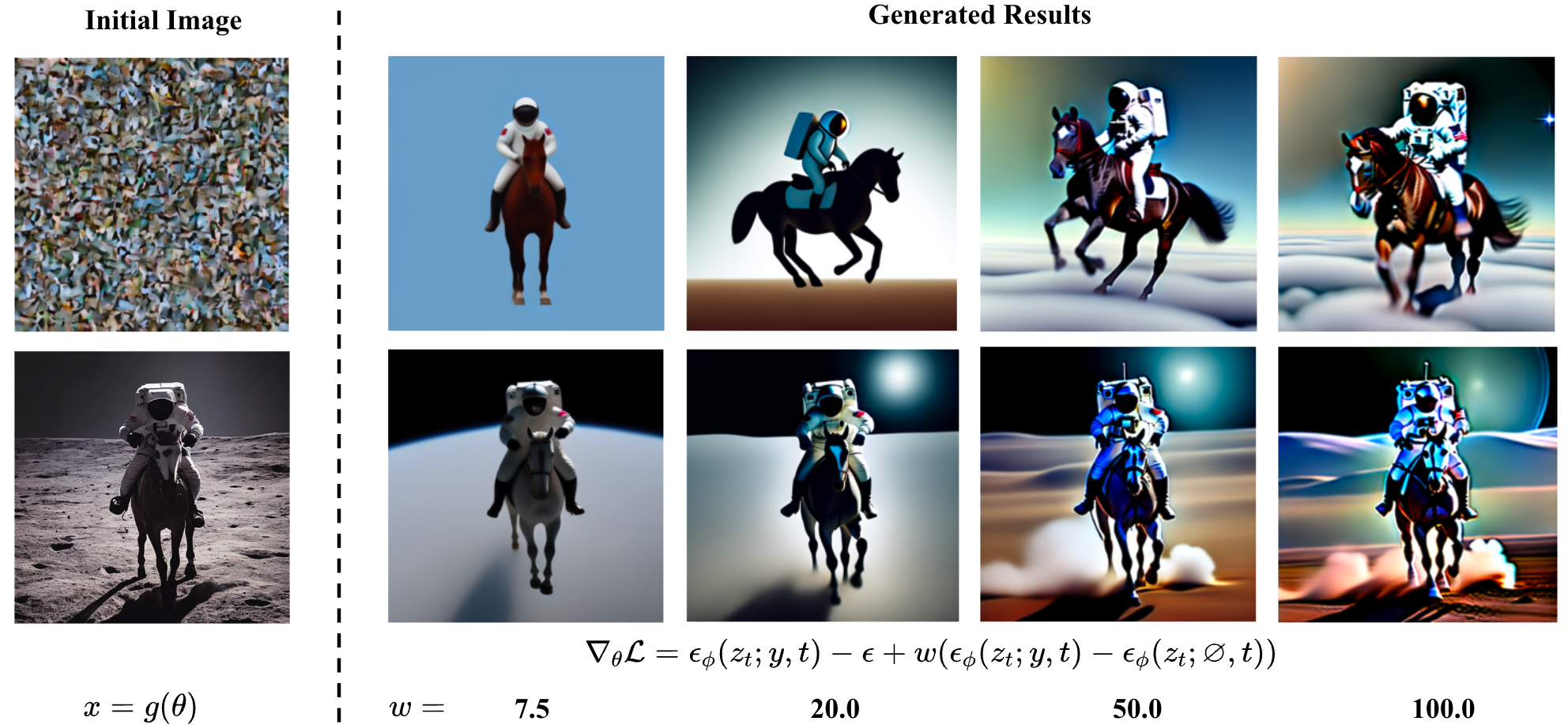}
    \caption{
        In order to show the relationship between over-saturation and guidance scale $w$ in SDS, we use different guidance scales for text-to-image generation. We find that as $w$ increases, the over-saturation of the generated images gradually becomes serious, while as $w$ decreases, the images gradually become over-smoothing.
    }
    \label{fig:gs}
\end{figure}

As demonstrated in Fig. \ref{fig:items}, we find that although using $\epsilon_\phi(z_t; y, t)-\epsilon_\phi(z_t; \varnothing, t)$ alone can generate images aligned with the textual condition $y$, it suffers from serious over-saturation and distortion, this cannot be avoided even when using a high-quality image as initialization. To further investigate the relationship between over-saturation and the classifier-free guidance term, we conduct another experiment. As shown in Fig. \ref{fig:gs}, we find that as $w$ increases, the images generated by SDS gradually become full of details (such as the texture of the horse, the details of the astronaut), but the over-saturation problem is also gradually amplified. Therefore, we speculate that the over-saturation phenomenon in SDS comes from the larger guidance scale. This phenomenon is also reasonable because Stable Diffusion \cite{rombach2022high} has not used classifier-free guidance term alone for generating during pre-training. Therefore, this item itself cannot generate real samples, and amplifying its weight have a negative impact on generated results.

Additionally, we discover that using $\epsilon_\phi(z_t; y, t) - \epsilon$ alone is incapable of generating images. Even in cases where a high-quality image is used for initialization, using $\epsilon_\phi(z_t; y, t) - \epsilon$ alone reverts it back into a semantically meaningless noise image. Based on this phenomenon, we further analyze $\epsilon_\phi(z_t; y, t) - \epsilon$. According to the DDIM sampling \cite{song2020denoising}, we know diffusion models predict the $x_0^t$ at time step $t$ with the following formula:
\begin{equation}
    \label{eq:pred_x0}
    \begin{aligned}
       x_0^t=\frac{z_t-\sqrt{1-\alpha_t}\epsilon_\phi(z_t;y,t)}{\sqrt{\alpha_t}}
    \end{aligned}
\end{equation}
In the case of SDS, we know that $z_t$ is obtained by $z_t =\sqrt{\alpha_t}x+\sqrt{1 - \alpha_t}\epsilon$. Therefore, the Eq. \ref{eq:pred_x0} can be written as:
\begin{equation}
    \label{eq:sds_recon}
    \begin{aligned}
       x = x_0^t+\frac{\sqrt{1-\alpha_t}}{\sqrt{\alpha_t}}(\epsilon_\phi(z_t;y,t)-\epsilon)
    \end{aligned}
\end{equation}
In this form, we find that $\epsilon_\phi(z_t; y, t) - \epsilon$ is actually related to the $x_0^t$ predicted at time step $t$, meaning its function is to reconstruct $z_t$ back to its pre-noised state $x$. However, this presents two problems. One is the single-step reconstruction error caused by diffusion models \cite{salimans2022progressive,song2023consistency,liu2023instaflow}, they can not reconstruct the image well when $t$ is large. This is why using $\epsilon_\phi(z_t; y, t) - \epsilon$ alone causes an originally high-quality image to become noise (as shown in the third column of Fig. \ref{fig:items}). Another is that when the original image $x$ has noise, reconstructing it back to its original state still retains the noise (as shown in the third column in Fig. \ref{fig:denoise}). This directly leads to the following problems in SDS: at the beginning (when $x$ lacks semantic information), $\epsilon_\phi(z_t; y, t) - \epsilon$ inhibits $\epsilon_\phi(z_t; y, t)-\epsilon_\phi(z_t; \varnothing, t)$ from aligning $x$ with the text condition, and in the later stage of SDS (when $x$ already meets the text condition), $\epsilon_\phi(z_t; y, t) - \epsilon$ degrades $x$ into noise. Therefore, these two issues ultimately lead to over-smoothing and loss of detail in SDS optimization.

Based on the above analysis, we know that SDS assigns a large guidance scale to the classifier-free guidance term to mitigate the over-smoothing and details lost problems brought by the reconstruction term. However, this exacerbates the over-saturation problem brought by the classifier-free guidance term.

\subsection{Invariant Score Distillation}
\label{sec:isd}
To overcome the aforementioned problems, \ie, the over-smoothing and details lost problems brought by the reconstruction term and the over-saturation problem brought by the large classifier-free guidance scale. We introduce an invariant score term derived from the DDIM sampling process to replace the original reconstruction term. This term no longer reconstructs the noisy image to its original state in a single step but aligns the noisy image to the previous time step state, avoiding the error of single-step reconstruction. 

\begin{figure}[t]
    \centering
    \includegraphics[width=0.98\textwidth]{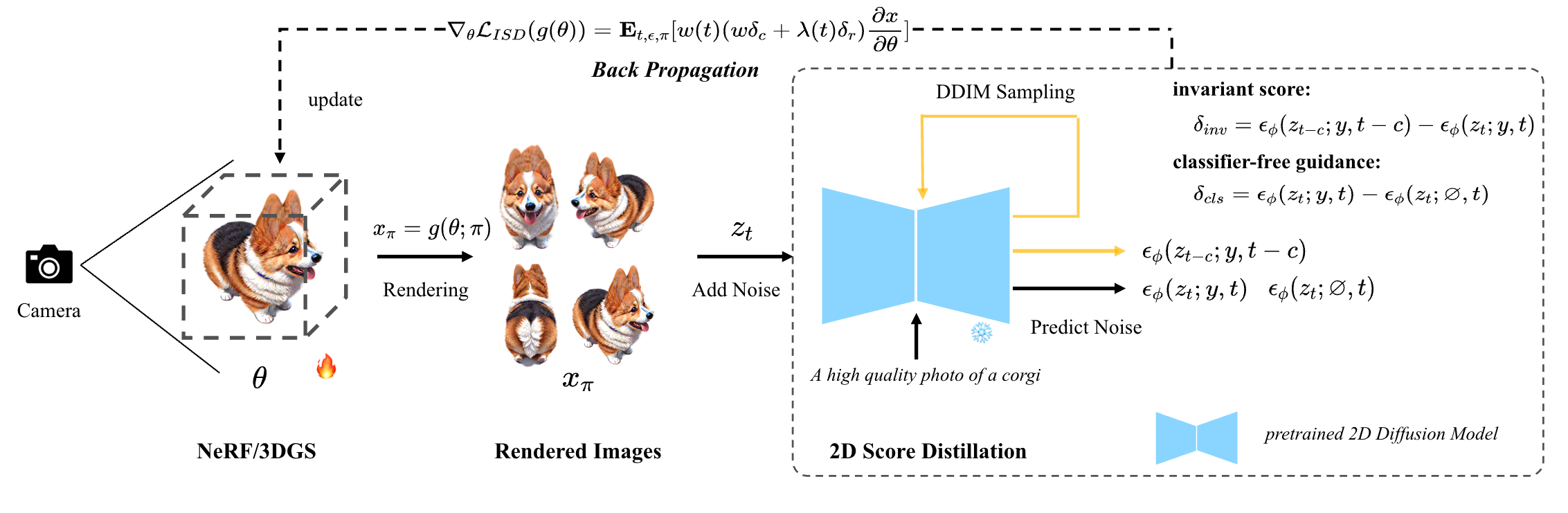}
    \caption{
        Overview of ISD for text-to-3D generation. We aim to optimize a 3D model $\theta$ using a pretrained text-to-image diffusion model. To achieve this, we render a 2D rendered image from $g(\theta, \pi)$ at a random pose $\pi$, and then employ a diffusion model $\epsilon_\phi$ to do the 2D Score Distillation. In particular, given a rendered image $x_\pi$, we first add noise to it to obtain $z_t$, and utilize the diffusion model to estimate the noise. In our framework, there are three noise predictions: $\epsilon_\phi(z_t, y, t)$, $\epsilon_\phi(z_t, \varnothing, t)$ and $\epsilon_\phi(z_{t-c}, y, t-c)$. We utilize them separately to compute the classifier-free guidance term $\delta_{cls}$ and the invariant score term $\delta_{inv}$, and then utilize our proposed ISD for optimization.
    }
    \label{fig:framework}
\end{figure}

According to the DDIM sampling formula, we can get the relationship between $x_0^t$ predicted at the current time step and $x_0^{t-c}$ predicted at the previous interval time step $t-c$:
\begin{equation}
    \label{eq:residual}
    \begin{aligned}
        x_0^{t-c} = x_0^t - \frac{\sqrt{{1-\alpha_{t-c}}}\epsilon_\phi(z_{t-c};t-c) - \sqrt{1-\alpha_{t-c}-\sigma_t^2}\epsilon_\phi(z_t;t)+\sigma_t\epsilon}{\sqrt{\alpha_{t-c}}}
    \end{aligned}
\end{equation}
We extract the interval sampling term about the noise prediction and get $\delta_{inv}=\epsilon_\phi(z_{t-c};y,t-c) - \sqrt{1-\sigma_t^2}\epsilon_\phi(z_t;y,t)-\sigma_t\epsilon$. Following the setting of DDIM, we set $\sigma_t=0$, and derive our invariant score term as follows:
\begin{equation}
    \label{eq:residual_term}
    \begin{aligned}
       \delta_{inv}=\epsilon_\phi(z_{t-c};y,t-c) - \epsilon_\phi(z_t;y,t)
    \end{aligned}
\end{equation}
where $c$ is the sampling step interval. Introducing this item brings two benefits: Firstly, by continuously aligning $x_0^t$ with the previous time step during the optimization process, the consistency of the final generated results is enhanced, thus overcoming the over-smoothing problems. Secondly, since this invariant score prior is related to $x_0^{t-c}-x_0^t$, and we know that $x_0^{t-c}$ predicted at the previous time step always has more details than $x_0^{t}$ at the current time step, so $x_0^{t-c}-x_0^t$ remains increased detail and removed the original noise (as shown in the last column of Fig. \ref{fig:denoise}), solving the original problem of details lost. Combing this term with the classifier-free guidance term $\delta_{cls}=\epsilon_\phi(z_t; y, t)-\epsilon_\phi(z_t; \varnothing, t)$, we obtain our final form of ISD, the gradient is expressed as follows:
\begin{equation}
    \label{eq:rsd}
    \begin{aligned}
       \nabla_\theta \mathcal{L}_{ISD}(g(\theta)) = \mathbf{E}_{t, \epsilon}[w(t)(\lambda(t)\delta_{inv}+w\delta_{cls})\frac{\partial x}{\partial \theta}]
    \end{aligned}
\end{equation}
where $\lambda(t)$ and $w(t)$ is weighting function, we keep the value of $w(t)$ consistent with SDS. In addition to the advantages discussed above, we found that ISD also allows us to assign a normal weight (7,5) to the classifier terms, thus avoiding the problem of over-saturation. In Fig. \ref{fig:framework}, we present an overview of our framework. Our framework can optimize a Neural Radiance Field (NeRF) \cite{mildenhall2020nerf,mueller2022instant} or 3D Gaussian Splatting \cite{kerbl3Dgaussians} using 2D text-to-image diffusion models. In Fig. \ref{fig:main_fig} and Fig. \ref{fig:3DGS}, we show our proposed ISD can generate high-fidelity and detailed 3D models in a single-stage optimization.
\subsection{Connection with other methods}
\label{sec:connect}
\begin{figure}[t]
    \centering
    \includegraphics[width=0.98\textwidth]{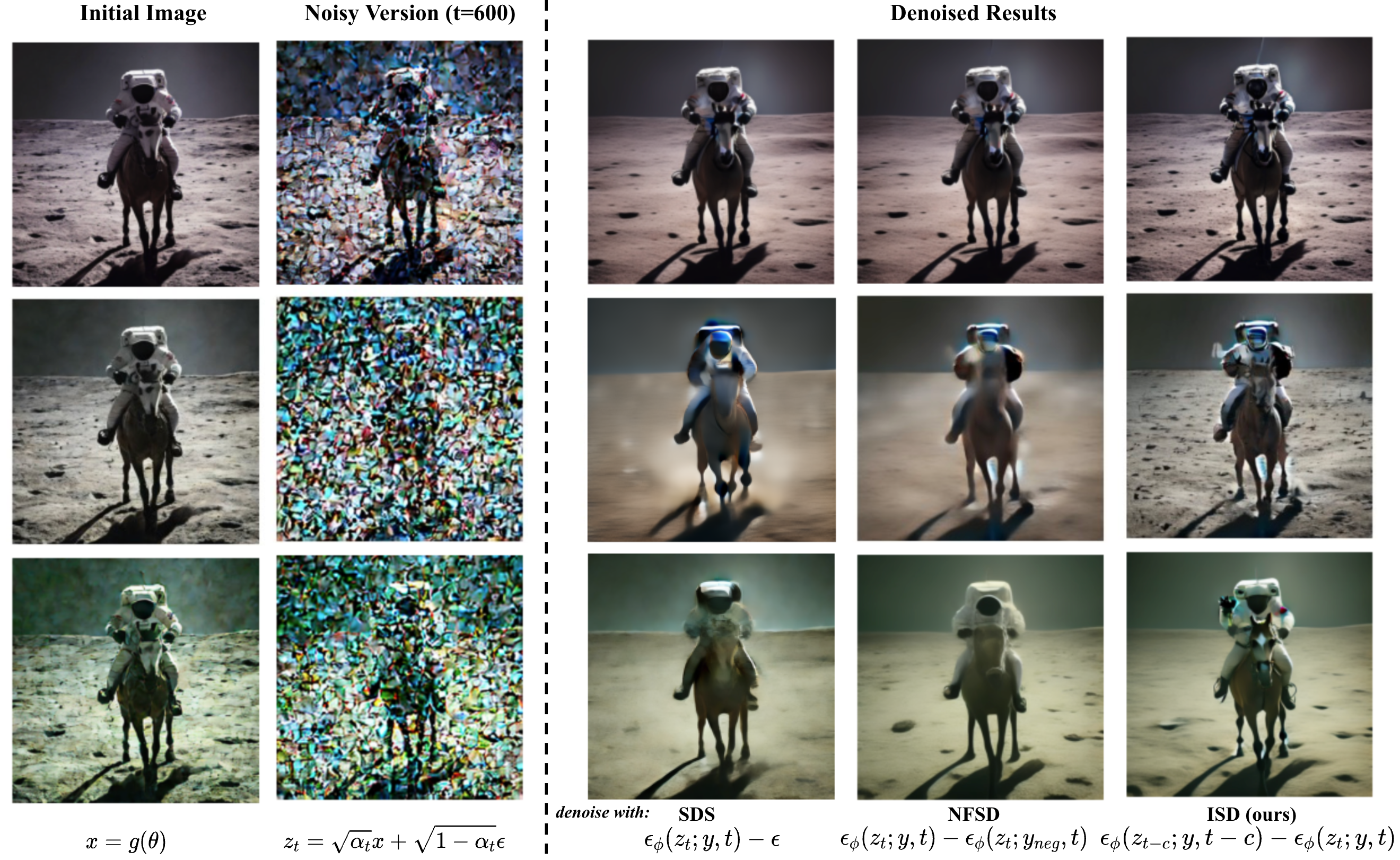}
    \caption{
        We use different noisy versions of a clear image as our initial images and add noise to each image with $t = 600$. We use different residual terms for denoising. The results show that when we continuously add noise to the initial image, our method can always restore more details than baseline methods
    }
    \label{fig:denoise}
\end{figure}
In this section, we discuss the connection of our proposed Invariant Score Distillation with several methods.

\noindent\textbf{Compared with SDS.} SDS \cite{poole2022dreamfusion} is proposed to distill 2D diffusion priors for text-to-3D generation. The objective function is given by Eq. \ref{eq:sds}. We find that it can be expressed as a special case of the residual term we introduced. Considering Eq. \ref{eq:residual} with $\sigma_t=\sqrt{1-\alpha_{t-1}}$, we observe that in this case, the formula reduces to:
\begin{equation}
    \label{eq:residual_sds}
    \begin{aligned}
        x_0^{t-1} = x_0^t + \frac{\sqrt{1-\alpha_{t-1}}}{\sqrt{\alpha_{t-1}}} (\epsilon-\epsilon_\phi(z_{t-1};t-1))
    \end{aligned}
\end{equation}
In this situation, we find that SDS is actually using the residual term $\epsilon_\phi(z_{t-1};t-1)-\epsilon$ to continuously optimize $x$. This is equivalent to using random noise $\epsilon \sim N(0, I)$ instead of $\epsilon_\phi(z_{t};t)$ to obtain $z_{t-1}$ after predicting $x_0^t$ at each step. This approach can cause significant interference in generating results.

\noindent\textbf{Compared with VSD.} VSD \cite{wang2024prolificdreamer} is proposed to alleviate the over-smoothing and over-saturated problems in SDS. In VSD, in addition to the original diffusion model $\epsilon_\phi(z_{t}; y,t)$, they also train an additional LoRA \cite{hu2021lora} $\epsilon_{lora}$. During the optimization process, they initialize the LoRA with the weights of $\epsilon_\phi(z_{t}; y,t)$ and fine-tuned with the rendered images from the current scene states. They replace the noise term $\epsilon$ in Eq. \ref{eq:sds} with $\epsilon_{lora}$. The gradient is expressed as follows:
\begin{equation}
    \label{eq:vsd}
    \begin{aligned}
       \nabla_\theta \mathcal{L}_{VSD} = \mathbf{E}_{t, \epsilon, \pi}[w(t)(\tilde{\epsilon}_\phi(z_t; y, t) - \epsilon_{lora}(z_{t}; y,t))\frac{\partial x}{\partial \theta}]
    \end{aligned}
\end{equation}
Actually, the residual term of VSD is $\epsilon_\phi(z_{t}; y,t)-\epsilon_{lora}(z_{t}; y,t)$. Since the LoRA model is initialized with the weights of $\epsilon_\phi(z_{t}; y,t)$ and uses random noise $\epsilon$ as the learning target. Therefore, $\epsilon_{lora}(z_{t}; y,t)$ can be written as a linear combination of $\epsilon_\phi(z_{t}; y,t)$ and $\epsilon$. \ie, $\epsilon_{lora}(z_{t}; y,t)=\sqrt{1-\sigma_t^2}\epsilon_\phi(z_{t}; y,t)+\sigma_t\epsilon$. This is still consistent with the general form of the residual term we provided in Eq. \ref{eq:residual}.

\noindent\textbf{Compared with NFSD.} NFSD \cite{katzir2023noise} decomposes the loss function into three terms and omits the denoising term $\delta_N=\epsilon_\phi(z_t;y,t)-\epsilon$. The final form is:
\begin{equation}
    \label{eq:NFSD}
    \begin{aligned}
       \nabla_\theta \mathcal{L}_{NFSD} = \mathbf{E}_{t, \epsilon, \pi}[w(t)(\delta_D + w\delta_{cls})\frac{\partial x}{\partial \theta}]
    \end{aligned}
\end{equation}
where $\delta_{cls}$ has the same form as ours, and $\delta_D$ is a piecewise function. When $t>200$, $\delta_D=\epsilon_\phi(z_{t}; \varnothing,t)-\epsilon_\phi(z_{t}; y_{neg},t)$. We take an item $\delta_{cls}=\epsilon_\phi(z_t; y, t) - \epsilon_\phi(z_t; \varnothing, t)$ and add it to $\delta_D$, which turns the original $w$ in Eq. \ref{eq:NFSD} to $w-1$. At this time, $\delta_D$ becomes $\delta_D = \epsilon_\phi(z_t; y, t) - \epsilon_\phi(z_t; y_{neg}, t)$. Since $y_{neg}$ describes the disordered and chaotic direction. Therefore we can regard $\epsilon_\phi(z_t; y_{neg}, t)$ as the noisy version of $\epsilon_\phi(z_t; y, t)$, \ie , $\epsilon_\phi(z_t; y_{neg}, t) \approx \epsilon_\phi(z_{t+c}; y, t+c)$.

To further verify the advantages of our method, we conduct another experiment. We use different noisy versions of an image to simulate $x$ in SDS. Then we imitate the process of SDS, add noise to these images, and restore them with different residual terms. As shown in Fig \ref{fig:denoise}, we verified that the image after SDS denoising has the problem of losing details and retaining noise, and the ISD we proposed can not only supplement the details, but also remove the noise originally covered by the image.

\section{Experiments}

\subsection{Implementation Details}

\label{sec:implemntation}
We implement Invariant Score Distillation (ISD) using the threestudio \cite{threestudio2023} framework. For all results, we set the diffusion step interval $c$ as 20 and $\lambda(t)$ as $\sqrt{(\beta_{t-c}/\alpha_{t-c})/(\beta_{t}/\alpha_t)}$. In addition, we sample time steps $t \sim \mathcal{U}(0.02, 0.98)$ at first 5000 steps, and after then, we follow the annealed time schedule proposed by ProlificDreamer \cite{wang2024prolificdreamer} and sample time steps $t \sim \mathcal{U}(0.02, 0.50)$.
For text-to-NeRF, we use Stable Diffusion 2-1-base \cite{rombach2022high} as our pretrained text-to-image diffusion model with guidance scale $w=7.5$. We set the rendering resolution as 512x512 and optimized all 3D models for 25000 iterations using AdamW \cite{loshchilov2019decoupled} optimizer with a learning rate of 0.01. We set the camera distance range as $[1.5, 2.0]$ and fovy range as $[40, 70]$, we found that these two parameters are highly related to the "Janus" problem. When set too low, it leads to a severe "Janus" problem. For text-to-3DGS, we set the rendering resolution as 512x512 and optimized all 3D models for 5000 iterations. We set the camera distance range as $[3.5, 5.0]$. For other training hyperparameters, we follow the original 3DGS \cite{kerbl3Dgaussians} paper. For the head generation, we add a description of hair in the prompt to avoid multi-face issues and utilize Point-E \cite{nichol2022point} as the 3D prior model with the prompt "a man head". We conduct all experiments on a single A6000 GPU.

\begin{figure}[!t]
    \centering
    \includegraphics[width=0.98\textwidth]{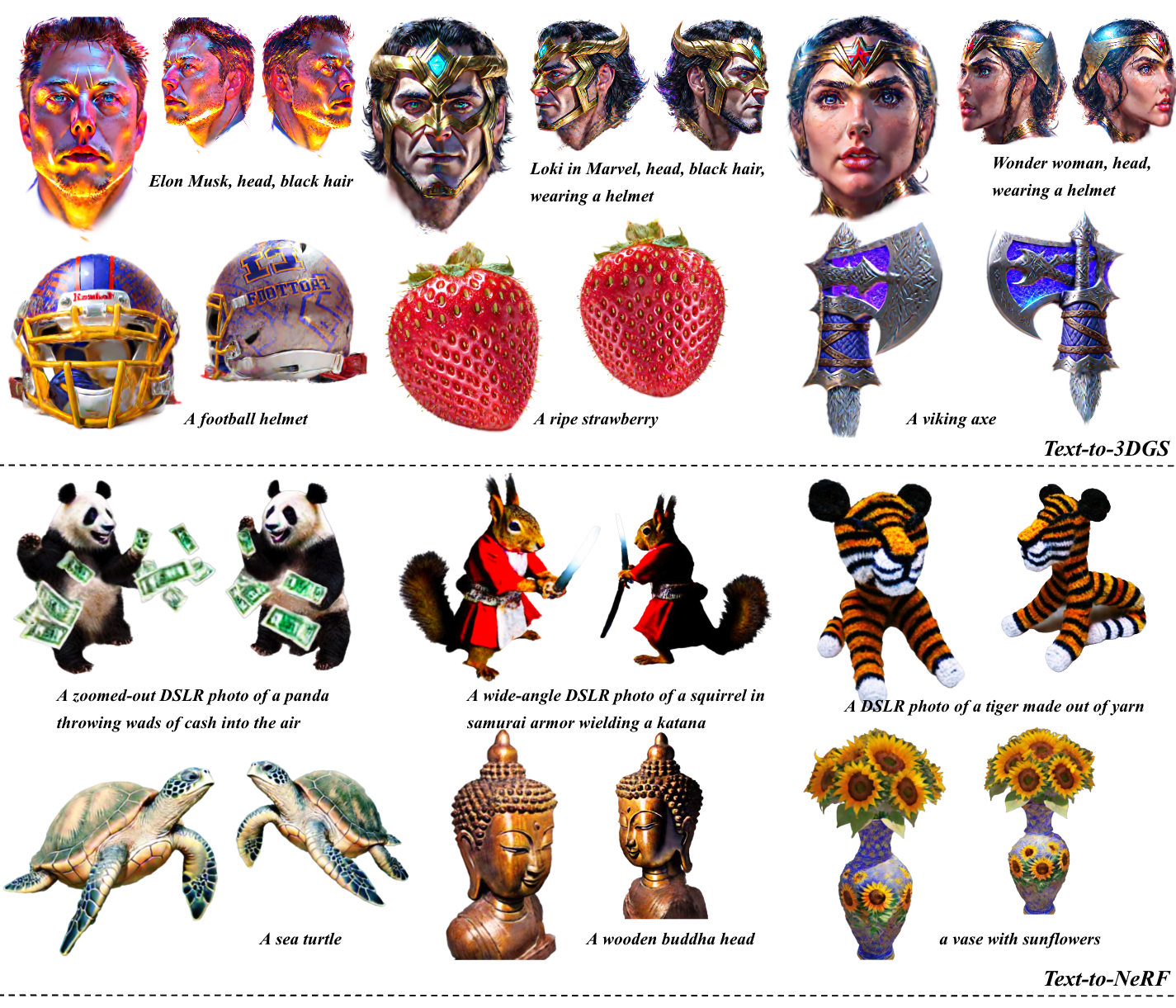}
    \caption{
        Examples generated by ISD. We provide examples of text-to-3DGS (the top two rows) and text-to-NeRF (the bottom two rows). ISD is capable of generating meticulously detailed and realistic textures, and it is also effective for long text descriptions.
    }
    \label{fig:3DGS}
\end{figure}

\subsection{Text-to-3D Generation}
\label{sec:t23d}

We present some representative results using ISD in Fig. \ref{fig:main_fig} and \ref{fig:3DGS}, including both text-to-3DGS and text-to-NeRF (bottom two lines in Fig. \ref{fig:3DGS}). In general, ISD can generate 3D assets aligned well with textual prompts and possess a realistic and detailed appearance. For text-to-3DGS, we show that ISD can generate highly intricate details. For text-to-NeRF, we also show that ISD can effectively generate long text descriptions of content.

\begin{figure}[!t]
    \centering
    \includegraphics[width=0.98\textwidth]{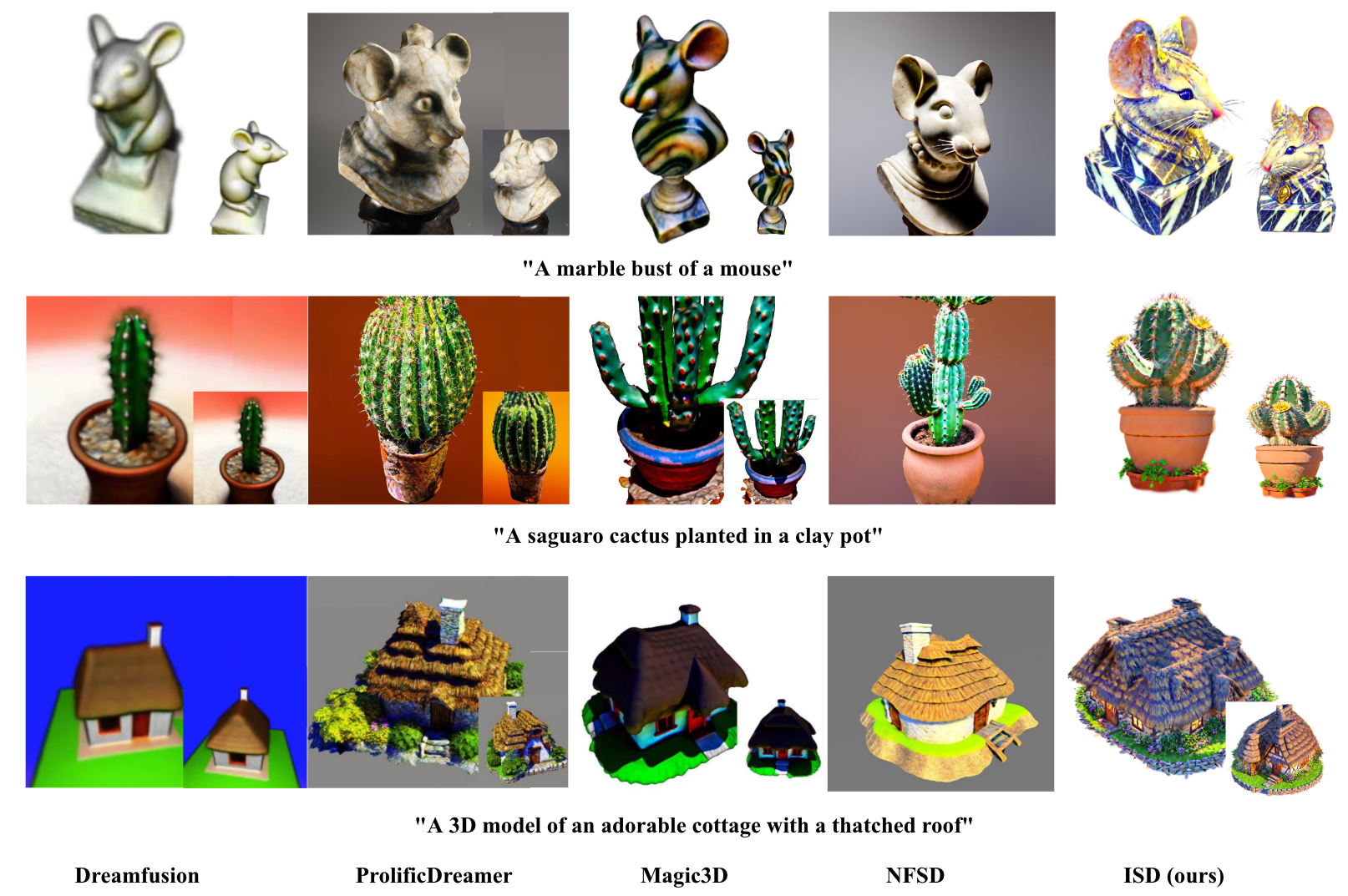}
    \caption{
        Qualitative comparisons to baselines for text-to-3D generation. Our method can generate 3D objects that align well with input text prompts with realistic and detailed appearances.
    }
    \label{fig:compare}
\end{figure}

\begin{figure}[!ht]
    \centering
    \includegraphics[width=0.98\textwidth]{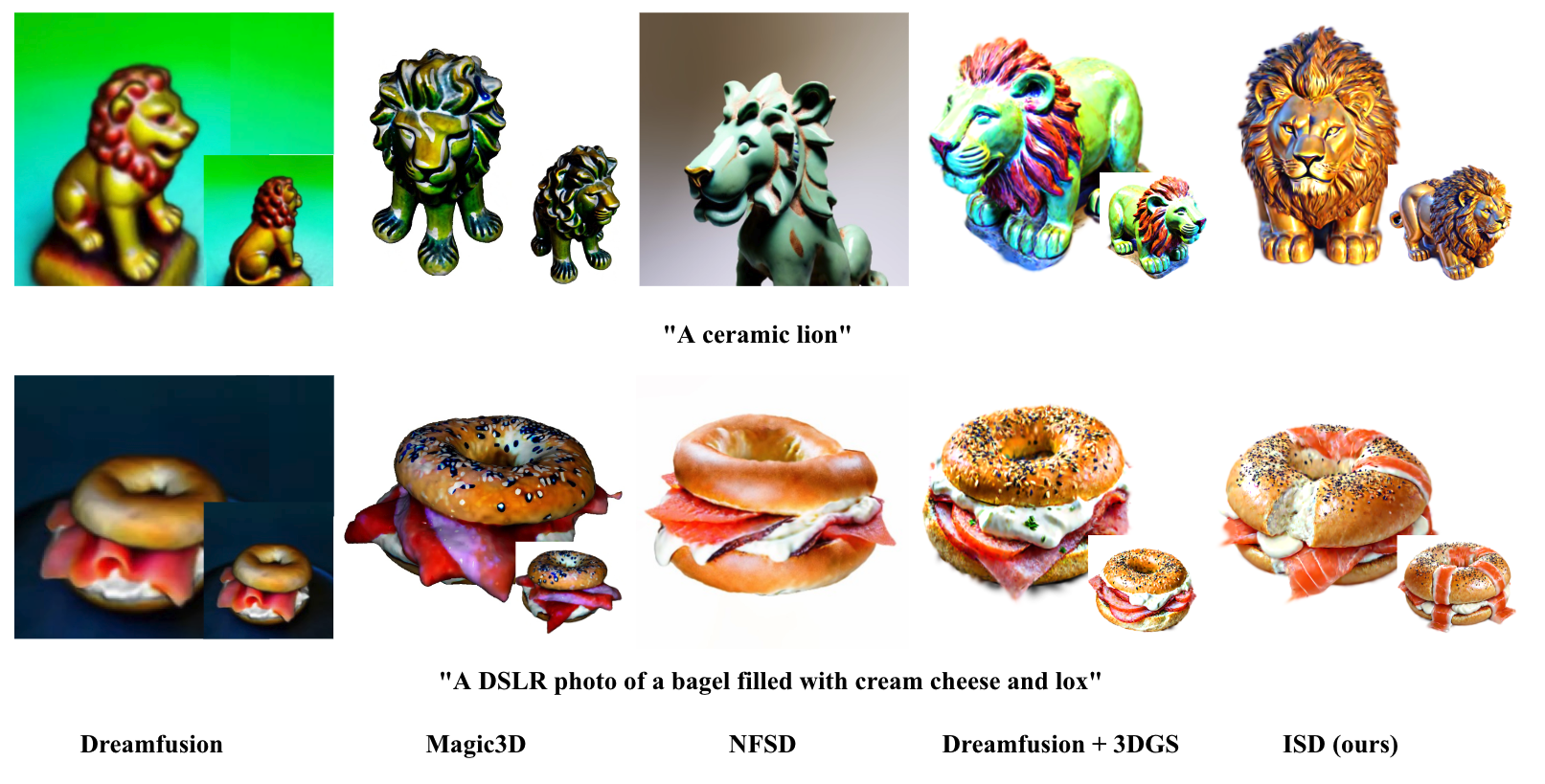}
    \caption{
        Qualitative comparisons to baselines for text-to-3D generation. We applied SDS and ISD to 3DGS for comparison. Although the quality generated by SDS has been greatly improved, ISD can still produce more detailed and realistic objects.
    }
    \label{fig:compare2}
\end{figure}

\noindent\textbf{Qualitative Comparisons.} In Fig. \ref{fig:compare}, we compare our generation results with Dreamfusion \cite{poole2022dreamfusion}, Maigc3D \cite{lin2023magic3d}, ProlificDreamer \cite{wang2024prolificdreamer} and NFSD \cite{katzir2023noise}. For DreamFusion, ProlificDreamer, and Magic3D, we take the results provided in ProlificDreamer. For NFSD, we directly use the results provided in their paper. As can be seen, our method achieves comparable or better results, while being significantly simpler than most of these methods. For example, observing the roof of the cottage, the results generated by our method have better detail and realism than other methods and are comparable to the result of ProlificDreamer. However, our method requires only minor changes and only single-stage optimization, compared to ProlificDreamer, we have great advantages in efficiency. Moreover, for a fair comparison, we apply ISD and SDS \cite{poole2022dreamfusion} to 3DGS \cite{kerbl3Dgaussians} technology. As can be seen in Fig. \ref{fig:compare2}, although the SDS results are much improved compared to the original version, our method can generate more realistic and detailed 3D objects. For example, in the bagel example, ISD produced a more detailed salmon texture.

\begin{table}[!t]
\begin{center}
\begin{minipage}{0.53\textwidth}
    \caption{User study results. We provide two attributes to measure the quality of 3D generation. Users choose the one they like most among the results we provide, the results show more users prefer our results.}
    \centering
    \resizebox{\textwidth}{!}{
    
    \begin{tabular}{lcc} 
    \toprule 
    Methods & Prompt Alignment $(\%)\uparrow$  &  Image Quality $(\%)\uparrow$ \\
    \midrule
    DreamFusion~\cite{poole2022dreamfusion}  & 10.74   & -  \\
    Magic3D~\cite{lin2023magic3d}  & 10.44  & 4.97  \\
    ProlificDreamer~\cite{wang2024prolificdreamer}  & 20.37  &  22.37  \\
    NFSD~\cite{katzir2023noise} & 17.53 & 10.82 \\
    Ours  & \textbf{40.92}   & \textbf{61.84} \\
    \bottomrule 
    \end{tabular}
    }
    \label{table:user1}
\end{minipage}
\hfill
\begin{minipage}{0.45\textwidth}
    \caption{Quantitative comparisons to baselines for text-to-3D generation, evaluated by CLIP Score and CLIP R-Precision.}
    \vspace{-1em}
    \centering
    \resizebox{\textwidth}{!}{
    \begin{tabular}{lcccc} 
    \toprule 
    Methods  & CLIP  & \multicolumn{3}{c}{CLIP R-Precision $(\%)\uparrow$} \\
               & Score $(\uparrow)$ & R@1 & R@5 & R@10 \\
    \midrule
    DreamFusion~\cite{poole2022dreamfusion}  & 67.22 & 75.79 & 91.70 & 97.42 \\
    Magic3D~\cite{lin2023magic3d}  & 75.73 & 77.31 & 92.74 & 96.93 \\
    ProlificDreamer~\cite{wang2024prolificdreamer}  & 78.90 & 78.64 & 93.44 & 97.42 \\
    NFSD~\cite{katzir2023noise} & 75.27 & 76.93 & 92.74 & 97.27 \\
    Ours  & \textbf{81.06} & \textbf{82.30} & \textbf{95.08} & \textbf{97.93} \\
    \bottomrule 
    \end{tabular}
    }
    \label{table:quantitative_results}
\end{minipage}
\end{center}
\end{table}
\noindent\textbf{Quantitative Evaluations.} Following previous works \cite{jain2021dreamfields,poole2022dreamfusion,luo2023scalable,yu2023text}, we quantitatively evaluate the quality of generated 3D assets utilizing CLIP Score \cite{hessel2022clipscore,radford2021learning} and CLIP R-Precision \cite{park2021benchmark}. We generate 3D objects using 50 distinct textual prompts. For each object, images are rendered from four different views (front, back, left, right). The resulting CLIP Score is calculated as the average semantic similarity between the textual prompt and the images from these four angles. We use the CLIP ViT-B/32 \cite{radford2021learning} model to extract text and image features, and the results are shown in Tab. \ref{table:quantitative_results}. Our approach significantly outperforms DreamFusion, Magic3D, ProlificDreamer, and NFSD in terms of CLIP Score and CLIP R-Precision, indicating better alignment between the generated results and input text prompts.

\noindent\textbf{User Study.} We follow previous works \cite{yu2023text,lin2023magic3d,wang2024prolificdreamer,katzir2023noise} and conducted a user study comparing rendered images obtained by Dreamfusion \cite{poole2022dreamfusion}, Magic3D \cite{lin2023magic3d}, ProlificDreamer \cite{wang2024prolificdreamer}, NFSD \cite{katzir2023noise} and our ISD method. We enlist 47 participants and ask them to choose the image that is most aligned with the given prompt, and the highest quality image. The result is presented in Tab. \ref{table:user1}, all methods provide results that are aligned with the prompt and our method attains the highest score. In terms of the rendered image quality, our method obtains significantly better scores compared with the competing methods.
\subsection{Ablation Study} 
\begin{figure}[t]
    \centering
    \includegraphics[width=0.98\textwidth]{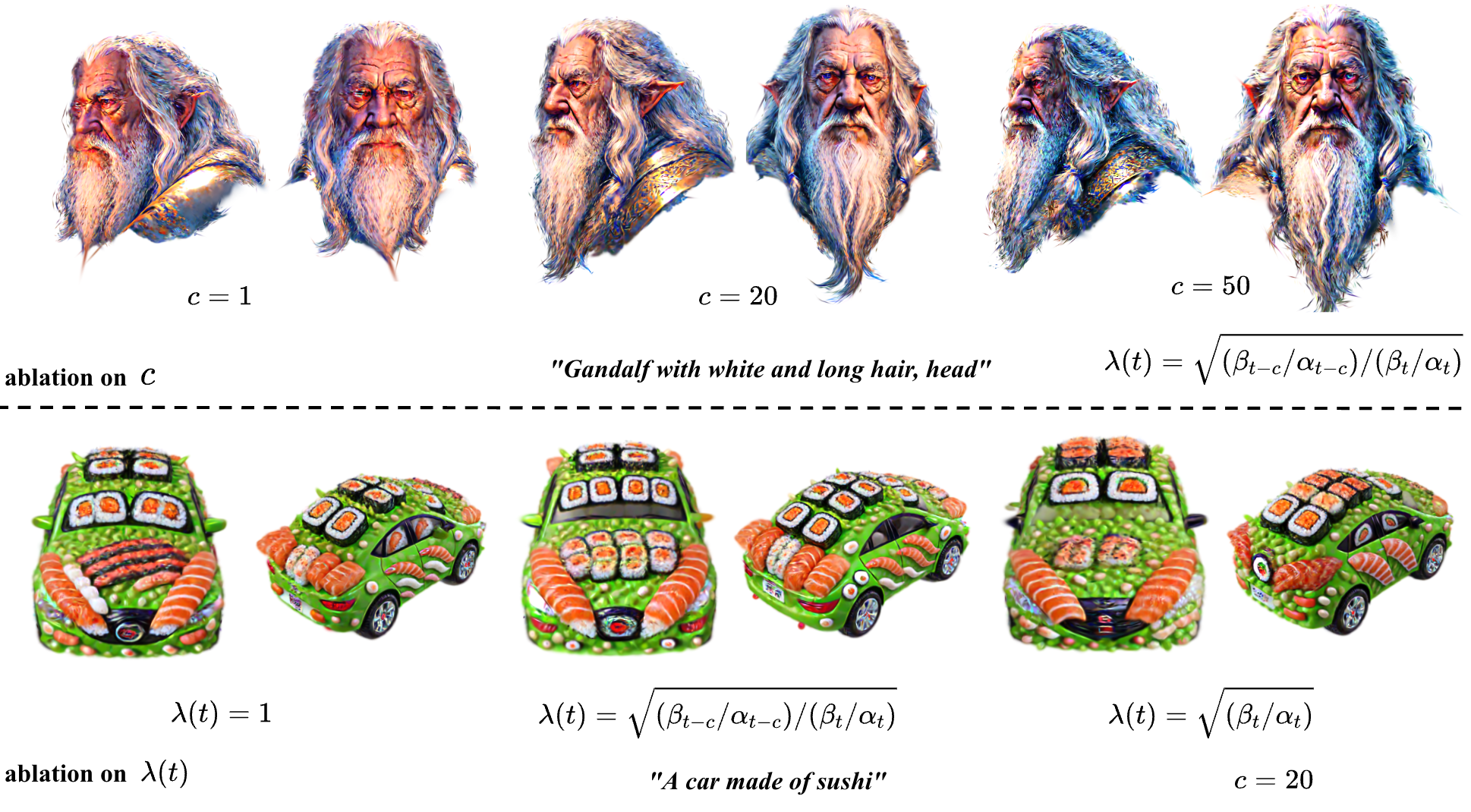}
    \caption{
        ISD with different $c$ (the first line) and $\lambda(t)$ (the second line). The results show that $c$ affects the details and quality of the generated results, and $\lambda(t)$ affects the richness of detail and alignment with the prompt.
    }
    \label{fig:ablation_c}
\end{figure}

\noindent\textbf{Ablation on time step interval $c$.} In the first line of Fig. \ref{fig:ablation_c}, we show the results generated using different $c$. We find that when $c$ is small (\eg, 1), the generated results lack details and suffer from over-saturation. And when $c$ is large (\eg, 50), the results have rich details, but there are problems such as blur artifacts. The reason here is that when $c$ gradually decreases, ISD gradually degenerates into only using the classifier-free guidance term to optimize the 3D model, and when $c$ is large, the increased variance will destroy the quality of the generated results and cause local over-smoothing. Therefore, we ultimately choose $c=20$ to avoid the problems of over-saturation and over-smoothing. The results show that when $c=20$, ISD can produce highly detailed and high-fidelity results.

\noindent\textbf{Ablation on weighting function $\lambda(t)$.} In the second line of Fig. \ref{fig:ablation_c}, we show the results generated using different $\lambda(t)$. We set $\lambda(t)$ according to SDS \cite{poole2022dreamfusion}, Magic3D \cite{lin2023magic3d}, and the derivation of Eq. 15. This coefficient determines the size of parameter weights at different time steps. The difference in parameter weights also has a certain impact on the generated results. As shown in Fig. \ref{fig:ablation_c}, we find that when $\lambda(t)=1$ and $\lambda(t)=\sqrt{(\beta_{t-c}/\alpha_{t-c})/(\beta_{t}/\alpha_t)}$, where $\beta_t=1-\alpha_t$, the sushi car is full of details, but we find that when $\lambda(t)=\sqrt{(\beta_{t-c}/\alpha_{t-c})/(\beta_{t}/\alpha_t)}$, the generated object is more consistent with the text description. 

\label{sec:ablation}

\section{Conclusion}
In this paper, we present an analysis of the over-saturation and over-smoothing problems in Score Distillation Sampling. We find that over-saturation comes from the larger classifier-free guidance scale, while over-smoothing comes from the noise preservation of the reconstruction term and the single-step reconstruction error. To overcome these problems, we propose Invariant Score Distillation (ISD), a novel
framework for text-to-3D generation. ISD replaces the reconstruction term with an invariant score term derived from DDIM sampling, and improves the problem of the reconstruction term in two aspects: 1) invariant score term no longer reconstructs the noisy but aligns the noisy image to its previous time step state, avoiding the single-step reconstruction error. 2) the invariant score term can remove the original noise in the image and add details, avoiding the details lost problem. Moreover, this improvement also allows us to use a normal classifier-free guidance scale, which avoids over-saturation problems. Extensive experiments demonstrate that our method greatly enhances SDS and produces realistic 3D objects through single-state optimization. 

\par\vfill\par

\section*{Acknowledgments}
This work is supported by the Major program of the National Natural Science Foundation of China (T2293720/T2293723), the National Natural Science Foundation of China (U2336212), the National Natural Science Foundation of China (62293554), the Fundamental Research Funds for the Central Universities (No. 226-2024-00058), the Fundamental Research Funds for the Zhejiang Provincial Universities (226-2024-00208), and the China Postdoctoral Science Foundation (524000-X92302).
%
%

\bibliographystyle{splncs04}
\bibliography{egbib}
\clearpage
\appendix
\section{Identify the problem in SDS}
\subsection{Implementation Details}
We present here the specific details of our text-to-image generation experiments to facilitate the reproduction of our experiments. We use threestudio \cite{threestudio2023} framework to implement our 2D experiments. For all results, we set the image resolution to $64\times64$ and use stable diffusion 2-1 base \cite{rombach2022high} as our base model. We set the maximum iteration step number to 1000, the minimum time step and the maximum time step to 20 and 980 respectively. We set the learning rate of the target to $3e-2$ (this is very important to reproduce our current results). We keep other parameters consistent with the original parameters in threestudio \cite{threestudio2023}.



\subsection{Visualization of Over-smoothing}
\begin{figure}[]
    \centering
    \includegraphics[width=0.85\textwidth]{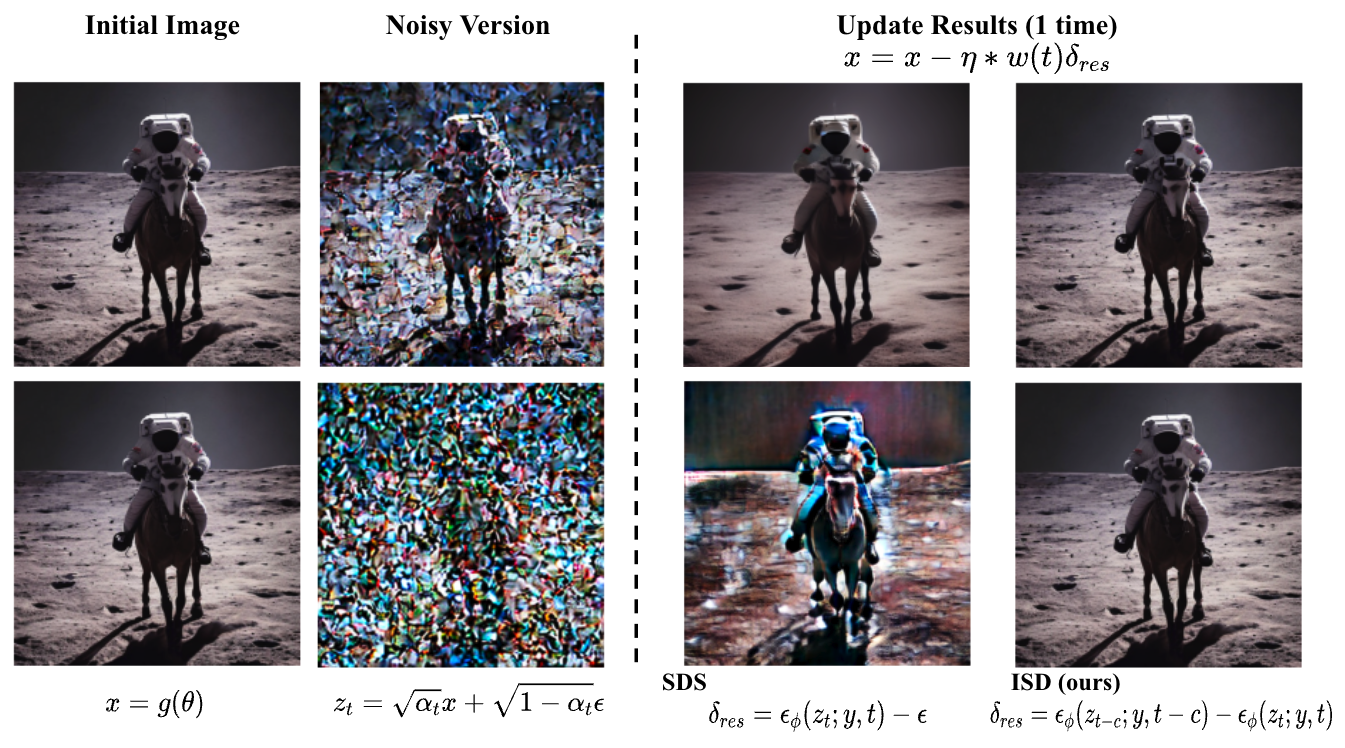}
    \caption{
        We use a high-quality image as initialization and use different t to add noise (t=300 in the first column and t=600 in the second column). We use different residual terms to update the image. The results show the residual term derived from SDS leads to serious over-smoothing.
    }
    \label{fig:update}
\end{figure}

In Sec. \ref{sec:deep_in_sds}, we briefly discussed the issue of the reconstruction term $\epsilon_\phi(z_t; y, t) - \epsilon$. Here, we show more specifically the problems caused by the original reconstruction term during the generation process. According Eq. \ref{eq:sds_recon}, we know that in SDS, the next update of $x$ can be seen as $x - \eta w(t) (\epsilon_\phi(z_t; y, t) - \epsilon)$, where $\eta$ is the learning rate. To better demonstrate the over-smoothing problem, we set $\eta$ to 1 and show the results after the image update. As shown in Fig. \ref{fig:update}, we find that even if we use a high-quality image as the initial image, after updating the image using SDS, the image becomes over-smoothing. Although in the actual process, $\eta$ is a very small value (\ie, $1e-5$), due to a large number of training steps, the accumulation of errors will still lead to over-smoothing. However, it is worth noting that using our proposed ISD, the original image will not be destroyed even in the presence of large noise.

\subsection{SDS fail in a normal guidance scale}
\begin{figure}[]
\centering
    \begin{minipage}{0.47\textwidth}
        \centering
        \includegraphics[width=1\textwidth]{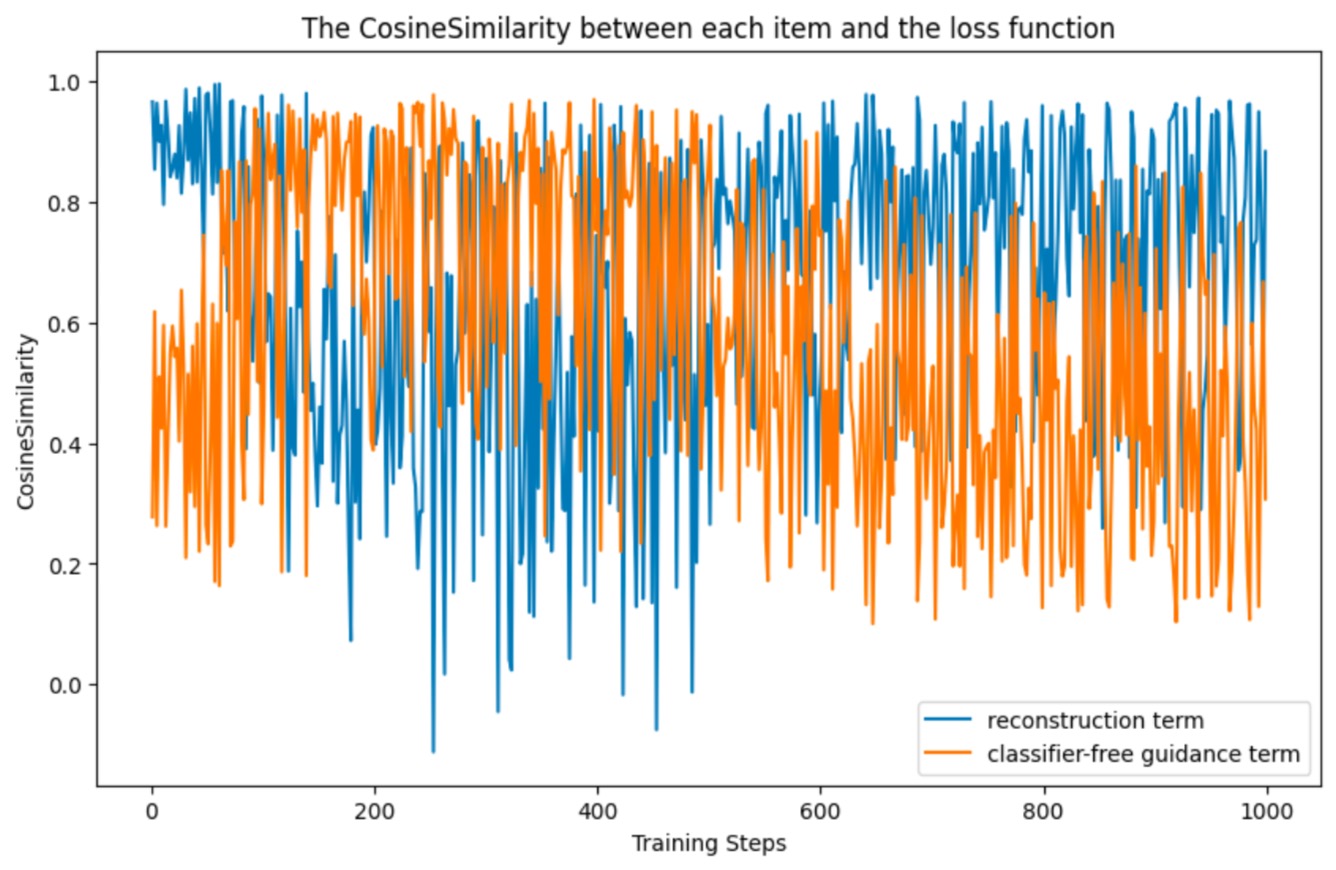}
        \caption{
            The cosine similarity between each item and the update loss in the SDS training process, with a classifier-free guidance scale=7.5. The higher the similarity, the greater its weight in the loss.
        }
        \label{fig:vis_sds}
    \end{minipage}
    \begin{minipage}{0.47\textwidth}
        \centering
        \includegraphics[width=1\textwidth]{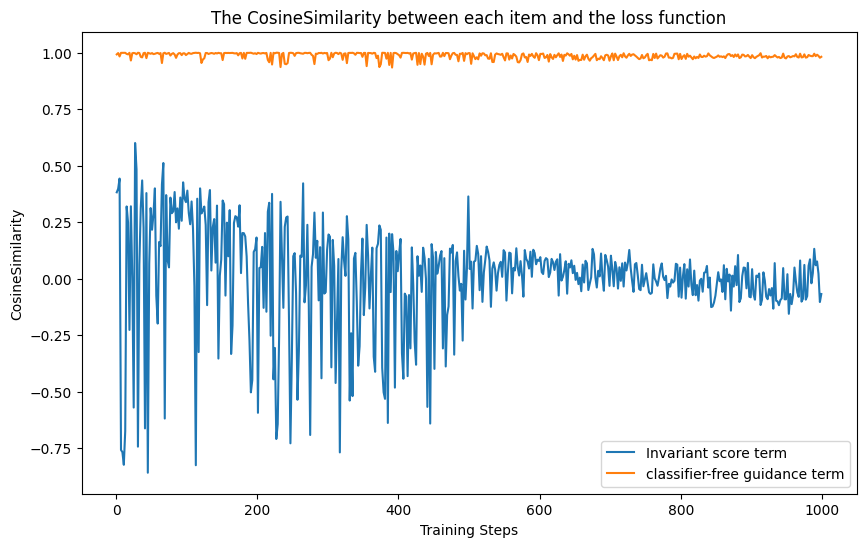}
        \caption{
            The cosine similarity between each item and the update loss in the ISD training process, with a classifier-free guidance scale=7.5. The higher the similarity, the greater its weight in the loss.
        }
        \label{fig:vis_isd}
    \end{minipage}
\end{figure}

In the previous content, we analyze that the reconstruction term caused over-smoothing and loss of details, while the larger guidance scale caused over-saturation problems. However, we found that using a small guidance scale produced results that lacked detail and were unrealistic (as shown in Fig. \ref{fig:gs}). This shows that if we want to solve the over-saturation problem, we must first solve the error caused by the reconstruction term. To further analyze the problem, we calculate the similarity of each item with the update loss. As shown in Fig. \ref{fig:vis_sds}, we find that at the beginning of the SDS training process, the reconstruction term is highly similar to the loss function. However, we know that the role of the classifier-free guidance term is to continuously guide the sample to align with the textual conditions, while the role of the reconstruction term is to restore the sample to its original state. Therefore, this shows that when the sample does not have semantics at the beginning of training, the reconstruction term restores the sample, hindering semantic alignment. At the end of the training, when the sample has semantic meaning, the final result is distorted due to the over-smoothing problem caused by the reconstruction term.

\begin{table}[!h]
\begin{minipage}[t]{0.38\textwidth}
    \caption{More evaluation}
    \centering
    \scalebox{1}{
    \resizebox{\textwidth}{!}{
        \begin{tabular}{lcc}
        \toprule
        Methods       & Average  & Average \\
                    & {saturation} & {brightness} \\
        \midrule 
        SDS & 203.75 & 171.74  \\
        NFSD & 170.60 & 190.14 \\
        VSD & 153.51 & 207.91 \\
        ISD & 144.92 & 216.97 \\

        \bottomrule
        \end{tabular}
    }}
    \label{table:more_eva}
\end{minipage} \begin{minipage}[t]{0.54\textwidth}
    \caption{More ablation on $c$}
    \centering
    \scalebox{1}{
    \resizebox{\textwidth}{!}{
    \begin{tabular}{lcccc} 
    \toprule 
    Methods  & CLIP  & \multicolumn{3}{c}{CLIP R-Precision $(\%)\uparrow$} \\
               & Score $(\uparrow)$ & R@1 & R@5 & R@10 \\
    \midrule
    ISD (c=0) & 73.37 & 75.69 & 90.07 & 96.53 \\
    ISD (c=1) & 76.59 & 79.93 & 93.35 & 97.42 \\
    ISD (c=20) & \textbf{81.06} & 82.30 & \textbf{95.08} & \textbf{97.93} \\
    ISD (c=50) & 80.87 & \textbf{82.44} & 94.98 & 97.79\\
    \bottomrule 
    \end{tabular}
    }}
    \label{table:quantitative_results}
\end{minipage}
\end{table}
\subsection{Quantify the Over-saturation}
To compare the saturation among methods, we calculated the \textbf{saturation} and \textbf{brightness} of 50 generated samples(Tab.~\ref{table:more_eva}). The average saturation of ISD is \textbf{29\%} (203.75 to 144.92) lower than that of SDS. In contrast, the average brightness of ISD is higher than that of SDS. This further proves the effectiveness and superiority of the ISD.

\subsection{Quantitative ablation study on hyperparameter $c$}
The hyperparameter $c$ is related to the richness of detail in the generated results.  We provide more quantitative results in Tab.~\ref{table:quantitative_results}. We find that a smaller $c$ results in a lower CLIP Score due to the loss of details, whereas a conventional value of $c$ yields better results.

\section{What makes ISD works}
\subsection{What makes ISD works}
ISD utilizes an invariant score term to replace the original reconstruction term, which not only avoids the loss of details and over-smoothing problems introduced by the original reconstruction term but also allows us to assign a smaller (7.5) weight to the classifier-free guidance term, avoiding the over-saturation problem. In Fig. \ref{fig:vis_isd}, we show the cosine similarity of each item with the update loss in the ISD training process. We find that the classifier-free guidance term always guides the alignment of samples with text conditions. Different from SDS, we find that the invariant score term introduced by ISD has a negative similarity to loss at the beginning, and fluctuates greatly. As the training progresses, the invariant score term gradually approaches 0. We explain that the rendered image is an out-of-domain sample at the beginning time, and the difference between different time steps is very large. Since the invariant score term continues to align the sample with its previous moment, this alignment not only adds more details to the sample but also ensures the consistency of the generation process. That's why our final results are realistic and full of detail.
\subsection{Discussion with NFSD}
In NFSD \cite{katzir2023noise}, the author split the SDS into three items and finally wrote the loss function as:
\begin{equation}
    \label{eq:nfsd_SDS}
    \begin{aligned}
       \nabla_\theta \mathcal{L}_{SDS} = \mathbf{E}_{t, \epsilon, \pi}[w(t)(\delta_D + \delta_N + w\delta_{C}-\epsilon)\frac{\partial x}{\partial \theta}]
    \end{aligned}
\end{equation}
where $\delta_C$ is equal to our $\delta_{cls}$. Then the authors experimentally find that the $\delta_N-\epsilon$ term introduced a bias, so they remove this term and obtain the final form of NFSD. When authors derive $\delta_D$ through experiments, they use a high-quality image as $x_{ID}$ and a noisy image as $x_{OOD}$, and the final $\delta_D$ is represented by the difference between the noise predictions corresponding to the two images, \ie, $\epsilon_\phi^{ID}-\epsilon_\phi^{OOD}$. This is consistent with the invariant score term we derived because $z_{t-c}$ can be seen as a clearer image than $z_t$, so we can approximately get $\epsilon_\phi^{ID}-\epsilon_\phi^{OOD} \sim \epsilon_\phi(z_{t-c};t-c) - \epsilon_\phi(z_t;t)$.

\section{Derivation of the invariant score term}
Here we more specifically derive how to get our invariant score term from the DDIM \cite{song2020denoising} sampling formula. Given a initial gaussian noise $z_t \sim N(0, I)$, DDIM first predicts the $x_0^t$ at the current time step $t$ through the following formula:
\begin{equation}
    \label{eq:pred_x0_2}
    \begin{aligned}
       x_0^t=\frac{z_t-\sqrt{1-\alpha_t}\epsilon_\phi(z_t;t)}{\sqrt{\alpha_t}}
    \end{aligned}
\end{equation}
then, it generates the sample $z_{t-1}$ via:
\begin{equation}
    \label{eq:denosing}
    \begin{aligned}
       z_{t-1} = \sqrt{\alpha_{t-1}}x_0^t + \sqrt{1-\alpha_{t-1}-\sigma_t^2}\epsilon_\phi(z_t;t)+\sigma_t\epsilon
    \end{aligned}
\end{equation}
where $\epsilon \sim N(0, I)$ is standard Gaussian noise independent of $x_t$. We consider Eq. \ref{eq:pred_x0_2} in the case of $t-1$ and replace $z_{t-1}$ with Eq. \ref{eq:denosing}, we can get:
\begin{equation}
    \label{eq:relation}
    \begin{aligned}
       x_0^{t-1} &= \frac{\sqrt{\alpha_{t-1}}x_0^t+\sqrt{1 - \alpha_{t-1}-\sigma_t^2}\epsilon_\phi(z_t;t)+\sigma_t\epsilon-\sqrt{1-\alpha_{t-1}}\epsilon_\phi(z_{t-1};t-1)}{\sqrt{\alpha_{t-1}}} \\
       &=x_0^t + \frac{\sqrt{1 - \alpha_{t-1}-\sigma_t^2}\epsilon_\phi(z_t;t)+\sigma_t\epsilon-\sqrt{1-\alpha_{t-1}}\epsilon_\phi(z_{t-1};t-1)}{\sqrt{\alpha_{t-1}}}
    \end{aligned}
\end{equation}
According to this formula, we find that during the 2D sampling process, we essentially use a residual term to optimize $x_0^t$ predicted at time step $t$ continuously. We extract this residual term as:

\begin{equation}
    \label{eq:residual_t}
    \begin{aligned}
        \delta_{res} = \epsilon_\phi(z_{t-1};t-1) - \sqrt{1-\sigma_t^2}\epsilon_\phi(z_t;t)-\sigma_t\epsilon
    \end{aligned}
\end{equation}
We can see that DDIM continuously supplements the image information by subtracting $\lambda(t)\delta_{res}$ from $x_0^t$, where $\lambda(t)$ is a weighting function. We make two changes to $\delta_{res}$ to get our final invariant score term. Firstly, we follow the setting of DDIM and set $\sigma_t=0$, indicating that we are removing the interference of random noise during the sampling process, making the sampling process a deterministic process. Secondly, due to the actual sampling process, we do not get $x_0^{t-1}$ from $x_0^t$ but use a time interval to directly update from $x_0^t$ to $x_0^{t-c}$, thus we change $t-1$ in Eq. \ref{eq:residual_t} to $t-c$, where $c$ is our time step interval. In our experiments, we set $c=20$ and demonstrate the effectiveness of this parameter through ablation experiments. Combine them, we get our final invariant score term as:
\begin{equation}
    \label{eq:ist}
    \begin{aligned}
        \delta_{inv} = \epsilon_\phi(z_{t-c};t-c) - \epsilon_\phi(z_t;t)
    \end{aligned}
\end{equation}
this is a special form of Eq. \ref{eq:residual_t}.
\begin{figure}[!ht]
    \centering
    \includegraphics[width=0.98\textwidth]{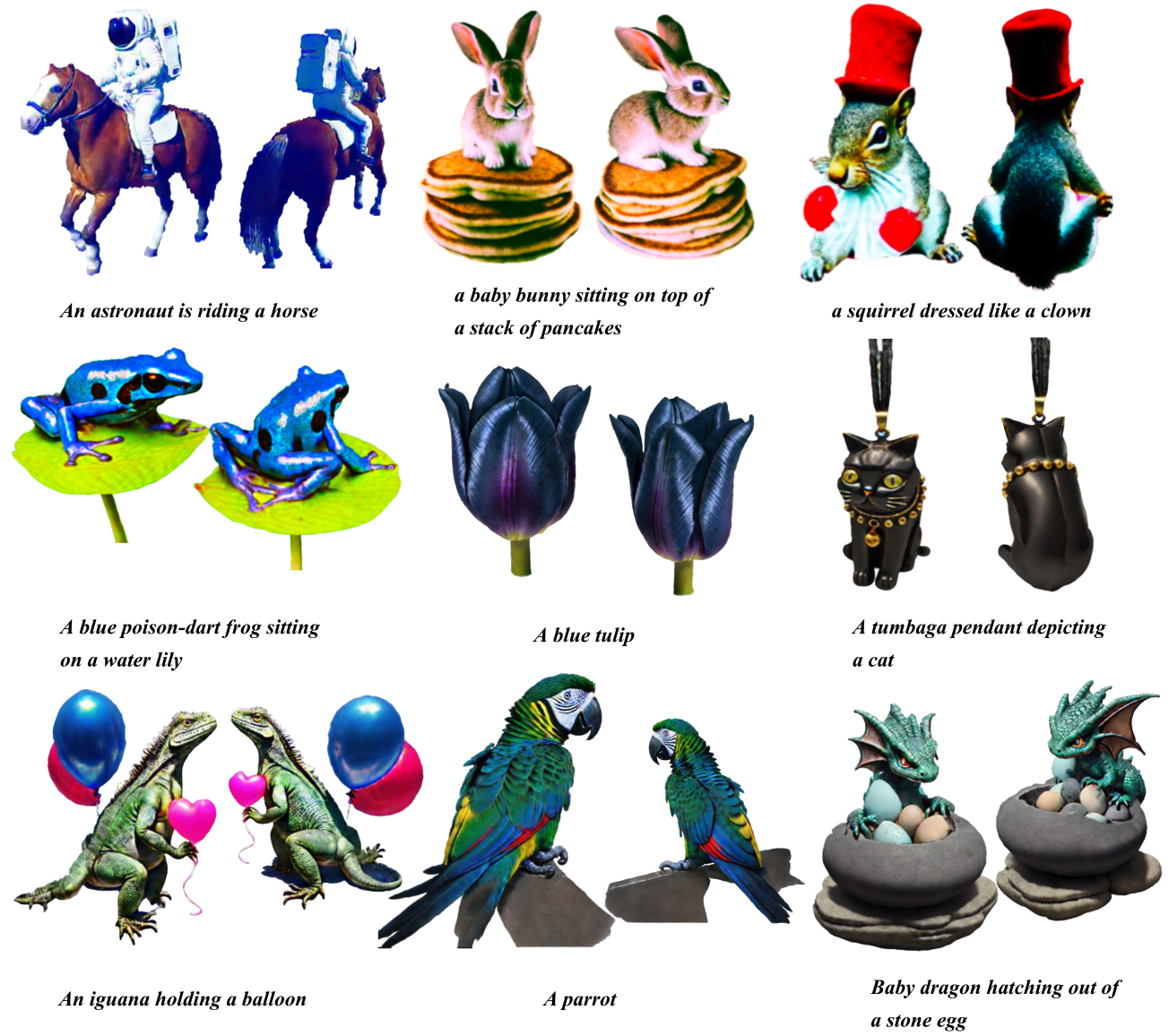}
    \caption{
        More text-to-NerRF examples generated by ISD.
    }
    \label{fig:more_nerf}
\end{figure}

\section{Additional Results}
In Fig. \ref{fig:more_nerf}, we present additional results for text-to-NeRF generation using our ISD.
\begin{figure}[!ht]
    \centering
    \includegraphics[width=0.98\textwidth]{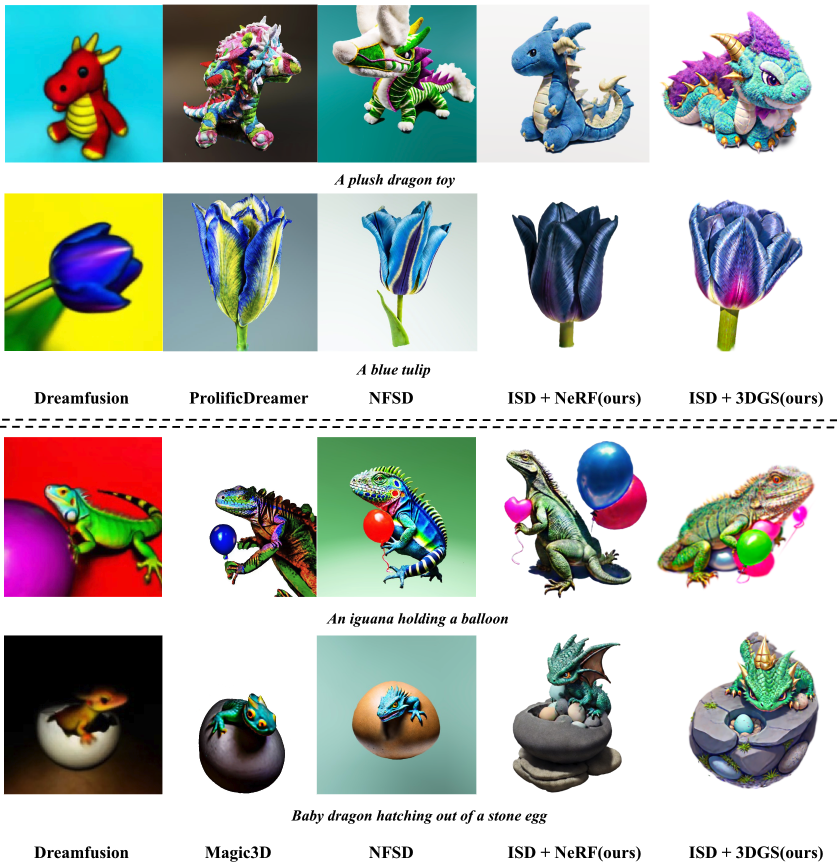}
    \caption{
        Comparisons of ISD with other methods.
    }
    \label{fig:more_comp2}
\end{figure}

\begin{figure}[!ht]
    \centering
    \includegraphics[width=0.98\textwidth]{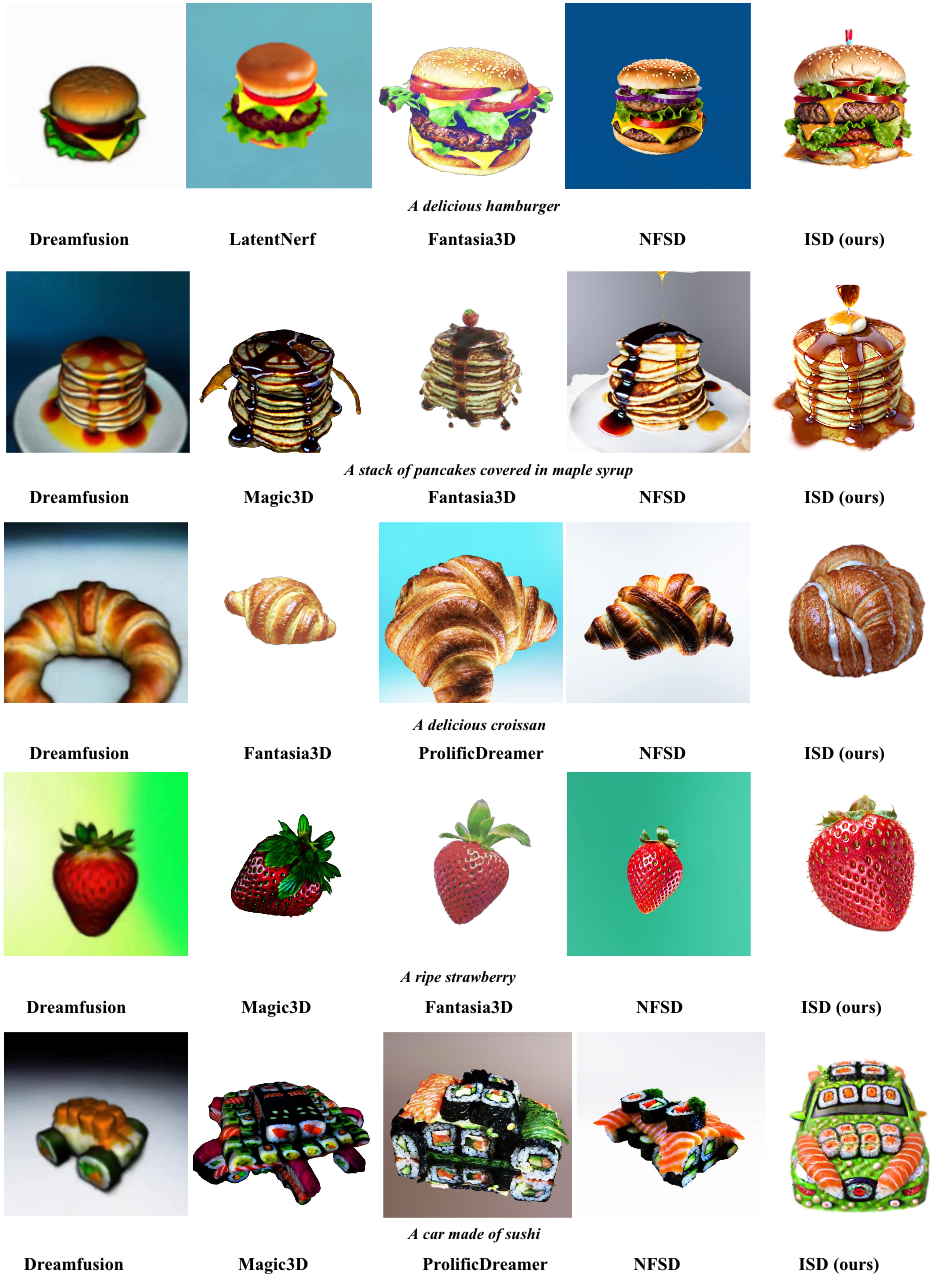}
    \caption{
        Comparisons of ISD with other methods.
    }
    \label{fig:more_comp}
\end{figure}

\begin{figure}[!ht]
    \centering
    \includegraphics[width=0.98\textwidth]{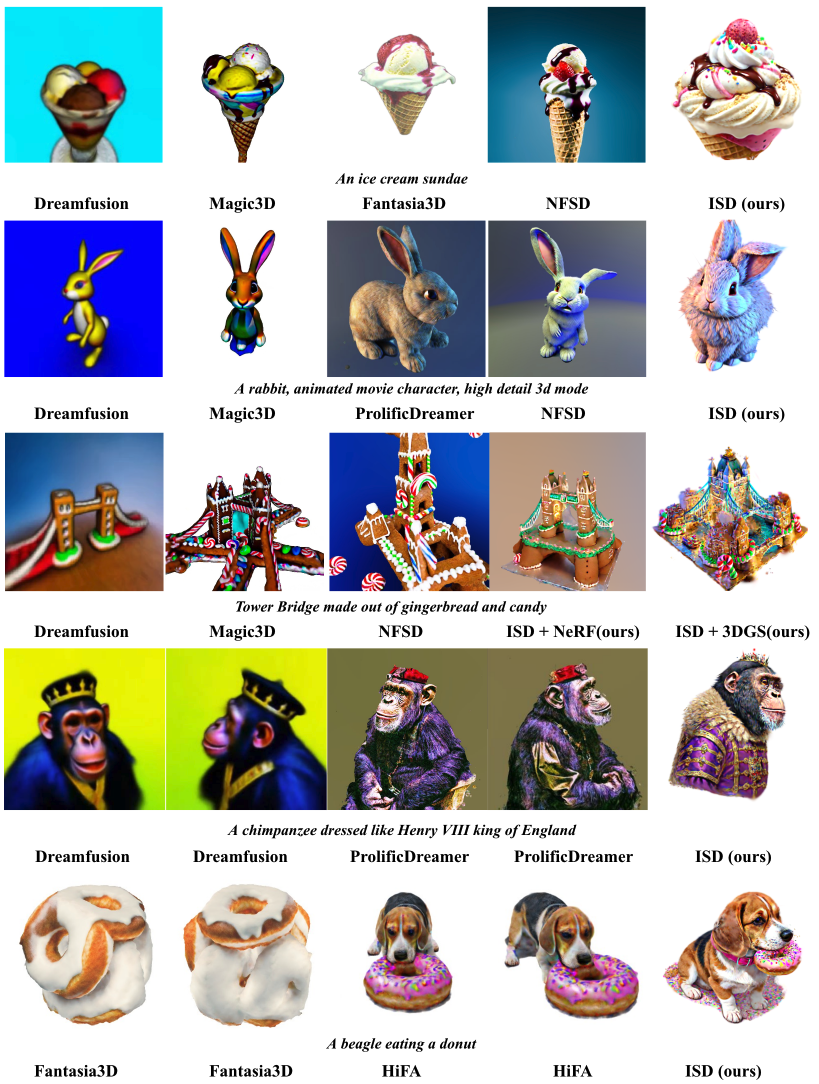}
    \caption{
        Comparisons of ISD with other methods.
    }
    \label{fig:more_comp3}
\end{figure}

Furthermore, in Fig. \ref{fig:more_comp2}, \ref{fig:more_comp}, \ref{fig:more_comp3}, we show additional comparisons of the results generated by our method with other methods.

\section{Limitation}
Invariant Score Distillation (ISD) mainly solves the problems of over-saturation and over-smoothing in SDS, but it does not specifically solve the Janus problem (multi-face), so the multi-face problem still exists in some examples. We find that two strategies can alleviate this problem: 1) Add descriptions about the back side in the prompt words (such as adding descriptions of hair in the human head example). 2) Increase the camera's field of view range (\ie, the fovy range). But we still need a method that can completely solve the multi-face problem, which will be our future exploration direction.
\end{document}